\documentclass{esannV2}

\usepackage{currfile}
\usepackage[mode=image|tex]{standalone}
\usepackage{xstring} 

\usepackage[T1]{fontenc}
\usepackage[utf8]{inputenc}
\usepackage{csquotes}

\usepackage[hypertexnames=false, unicode, pdfusetitle]{hyperref}

\usepackage{lmodern}
\usepackage{textcomp}
\usepackage{microtype}
\usepackage{etoolbox}
\robustify\bfseries

\usepackage[skip=3pt]{subcaption}

\usepackage{booktabs}
\usepackage{tabularx}
\usepackage{siunitx}

\usepackage{amsmath}
\usepackage{amssymb}
\usepackage{mathtools}
\usepackage{bm}
\usepackage{xfrac}

\usepackage{url}
\usepackage[capitalise, nameinlink]{cleveref}
\crefformat{equation}{(#2#1#3)}
\usepackage[style=numeric, backend=biber, url=false, maxcitenames=1]{biblatex}

\usepackage{todonotes}
\usepackage{blindtext}

\newcommand{\D}{\mathcal{D}}

\newcommand{\rv}[1]{\bm{#1}}

\newcommand{\mat}[1]{\bm{#1}}

\DeclareMathOperator{\softmax}{softmax}

\newcommand*{\diff}{\mathop{}\!\mathrm{d}}

\newcommand{\nth}[2][th]{#2^{\text{#1}}}

\DeclareMathOperator{\E}{\mathbb{E}}

\DeclareMathOperator{\p}{p}
\DeclareMathOperator{\q}{q}

\DeclareMathOperator{\Norm}{\mathcal{N}}
\DeclareMathOperator{\Multi}{\mathcal{M}}

\DeclareMathOperator{\Uni}{\mathcal{U}}
\DeclareMathOperator{\Ind}{\mathbb{I}}

\providecommand\given{}
\DeclarePairedDelimiterX{\Cond}[1]{(}{)}{
\renewcommand\given{%
  \nonscript\mkern2mu
  \delimsize\vert
  \nonscript\mkern2mu
  \mathopen{}
  \allowbreak}
#1
}

\makeatletter
\newcommand{\Fun}{\@ifstar\@sfun\@fun}
\newcommand{\@fun}[1]{#1\Cond}
\newcommand{\@sfun}[1]{#1\Cond*}
\makeatother

\newcommand{\Prob}{\p\Cond}

\newcommand{\Variat}{\q\Cond}
\newcommand{\Gaussian}{\Norm\Cond}
\newcommand{\Multinomial}{\Multi\Cond}

\DeclarePairedDelimiterX{\KLdelim}[2]{(}{)}{%
  #1\mkern2mu\delimsize\|\mkern2mu#2%
}

\DeclarePairedDelimiterXPP{\Moment}[2]{#1}{[}{]}{}{
\renewcommand\given{%
  \nonscript\mkern2mu
  \delimsize\vert
  \nonscript\mkern2mu
  \mathopen{}
  \allowbreak}
#2
}

\DeclarePairedDelimiterX{\Set}[1]{\{}{\}}{

#1
}

\DeclarePairedDelimiterXPP{\pix}[1]{\begingroup\scriptscriptstyle}{(}{)}{\endgroup}{\mkern-2mu#1\mkern-2mu}

\definecolor{tumblue}{HTML}{0065BD}
\definecolor{tumgreen}{HTML}{A2AD00}
\definecolor{tumorange}{HTML}{E37222}
\definecolor{tumivory}{HTML}{DAD7CB}
\definecolor{tumred}{HTML}{E53418} 
\definecolor{tumviolet}{HTML}{69085A} 

\definecolor{tumgray0}{HTML}{000000}
\definecolor{tumgray1}{HTML}{58585A}
\definecolor{tumgray2}{HTML}{9C9D9F}
\definecolor{tumgray3}{HTML}{D9DADB}
\definecolor{tumgray4}{HTML}{FFFFFF}

\definecolor{tumblue0}{HTML}{003359}
\definecolor{tumblue1}{HTML}{005293}
\definecolor{tumblue2}{HTML}{0073CF}
\definecolor{tumblue3}{HTML}{64A0C8}
\definecolor{tumblue4}{HTML}{98C6EA}

\definecolor{tumgreen0}{HTML}{EAF900}
\definecolor{tumgreen1}{HTML}{AEBA00}
\definecolor{tumgreen2}{HTML}{8A9300}
\definecolor{tumgreen3}{HTML}{525800}

\definecolor{tumred0}{HTML}{F23719}
\definecolor{tumred1}{HTML}{CB2E15}
\definecolor{tumred2}{HTML}{90210F}
\definecolor{tumred3}{HTML}{65170B}

\definecolor{tumorange0}{HTML}{F07824}
\definecolor{tumorange1}{HTML}{C9651E}
\definecolor{tumorange2}{HTML}{8E4715}
\definecolor{tumorange3}{HTML}{63320F}

\definecolor{sPetrol}{RGB}{0,153,153}
\definecolor{sStoneDark}{RGB}{60,70,75}
\definecolor{sStoneLight}{RGB}{135,155,170}
\definecolor{sStone}{RGB}{190,205,215}
\definecolor{sSandDark}{RGB}{115,100,90}
\definecolor{sSandLight}{RGB}{170,170,150}
\definecolor{sSand}{RGB}{215,215,205}
\definecolor{sSnow}{RGB}{255,255,255}

\definecolor{sTealDark}{RGB}{0,100,110}
\definecolor{sTealLight}{RGB}{65,170,170}
\definecolor{sBlueDark}{RGB}{0,95,135}
\definecolor{sBlueLight}{RGB}{80,190,215}
\definecolor{sGreenDark}{RGB}{100,125,45}
\definecolor{sGreenLight}{RGB}{170,180,20}
\definecolor{sYellowDark}{RGB}{235,120,10}
\definecolor{sYellowLight}{RGB}{255,185,0}
\definecolor{sRedDark}{RGB}{100,25,70}
\definecolor{sRedLight}{RGB}{175,35,95}

\definecolor{hannah0}{HTML}{d89d1c}
\definecolor{hannah1}{HTML}{d8801c}
\definecolor{hannah2}{HTML}{d8511c}
\definecolor{hannah3}{HTML}{d8bf1c}
\definecolor{hannah4}{HTML}{d8511c}

\begin{document}
\title{Interpretable Dynamics Models for Data-Efficient Reinforcement Learning}

\author{Markus Kaiser$^{1,2}$%
    \thanks{The project this report is based on was supported with funds from the German Federal Ministry of Education and Research under project number 01\,IS\,18049\,A.}
    \ and Clemens Otte$^1$
    and Thomas Runkler$^{1,2}$
    and Carl Henrik Ek$^3$
\vspace{.3cm}\\
1- Siemens AG, Germany.\quad 2- Technical University of Munich, Germany.\\
3- University of Bristol, United Kingdom.
}

\maketitle

\begin{abstract}
In this paper, we present a Bayesian view on model-based reinforcement learning.
We use expert knowledge to impose structure on the transition model and present an efficient learning scheme based on variational inference.
This scheme is applied to a heteroskedastic and bimodal benchmark problem on which we compare our results to NFQ and show how our approach yields human-interpretable insight about the underlying dynamics while also increasing data-efficiency.
\end{abstract}

\section{Introduction}
\label{sec:introduction}
In reinforcement learning (RL)~\cite{sutton_reinforcement_1998}, an agent's task is to learn a policy $\pi$ which, given the current state $\mat{s}$ of an environment, chooses an action $\mat{a}$ to achieve the goal specified by a reward function $r$ mapping states to numerical rewards.
The next state $\mat{s}^\prime = \Fun*{f}{\mat{s}, \mat{a}}$ is determined by the latent and possibly stochastic transition function $f$.
We consider batch RL problems~\cite{lange_batch_2012}, where we are presented with a set of state transitions $\D = \Set{(\mat{s}_n, \mat{a}_n, \mat{s}_n^\prime)}_{n=1}^N$ and are unable to interact with the original system to find a policy.
This setup is common in industrial applications of RL, where deploying an untrusted policy can lead to safety issues.
Similarly, gathering data can be costly, calling for data-efficient methods.

\Textcite{deisenroth_pilco_2011} showed how data-efficiency in model-based RL can be increased with probabilistic models for the transition dynamics $f$.
They provide a principled way of taking model uncertainty into account when evaluating the performance of a policy, thereby reducing the impact of model-bias.
A shortcoming of this approach is the limitations put on modelling choices by the inference scheme.
Transition dynamics are modelled as standard Gaussian processes (GPs) and policies and rewards must be of specific forms.
In this work, we extend this approach by allowing and imposing additional structure.
In many environments, experts can describe abstract properties of the system even if no closed form models are available.
Incorporating this knowledge facilitates learning and allows us to precisely state what we want to learn from data.

The paper is outlined as follows.
After introducing the heteroscedastic and bimodal Wet-Chicken benchmark, we show how high-level knowledge about this system can be used to impose Bayesian structure.
We derive an efficient inference scheme for both the dynamics model and for probabilistic policy search based on variational inference.
We show that this approach yields interpretable models and policies and is significantly more data-efficient than less interpretable alternatives.

\section{The Wet-Chicken Benchmark}
\label{sec:wetchicken}
In the Wet-Chicken problem~\cite{hans_efficient_2009}, a canoeist is paddling in a two-dimensional river.
The canoeist's position at time $t$ is given by $\mat{s}_t = (x_t, y_t)$, where $x_t$ denotes the position along the river and $y_t$ the position across it.
The river is bounded by its length $l = 5$ and width $w = 5$.
There is a waterfall at the end of the river at $x = l$.
The canoeist wants to get close to the waterfall to maximize the reward $\Fun*{r}{\mat{s}_t} = x_t$.
However, if the canoeist falls down the waterfall he has to start over at the initial position $(0, 0)$.

The river's flow consists of a deterministic velocity $v_t = y_t \cdot \sfrac{3}{w}$ and stochastic turbulence $b_t = 3.5 - v_t$, both of which depend on the position on the $y$-axis.
The higher $y_t$ the faster the river flows but also the less turbulent it becomes.
The canoeist chooses his paddle direction and intensity via an action $\mat{a}_t = (a_{t,x}, a_{t,y}) \in [-1, 1]^2$.
The transition function $f : (\mat{s}_t, \mat{a}_t) \mapsto \mat{s}_{t+1} = (x_{t+1}, y_{t+1})$ is given by
\begin{align}
    x_{t+1} &= \begin{cases}
        0 & \text{if } \hat{x}_{t+1} > l \\
        0 & \text{if } \hat{x}_{t+1} < 0 \\
        \hat{x}_{t+1} & \text{otherwise}
    \end{cases} &
    y_{t+1} &= \begin{cases}
        0 & \text{if } \hat{x}_{t+1} > l \text{ or } \hat{y}_{t+1} < 0 \\
        w & \text{if } \hat{y}_{t+1} > w \\
        \hat{y}_{t+1} & \text{otherwise}
    \end{cases}
\end{align}
where $\hat{x}_{t+1} = x_t + (1.5 \cdot a_{t, x} - 0.5)  + v_t + b_t \cdot \tau_t$ and $\hat{y}_{t+1} = y_t + a_{t, y}$ and $\tau_t \sim \Fun*{\Uni}{-1, 1}$ is a uniform random variable that represents the turbulence.

There is almost no turbulence at $y = w$, but the velocity is too high to paddle back.
Similarly, the velocity is zero at $y = 0$, but the canoeist can fall down the waterfall unpredictably due to the high turbulence.
A successful canoeist must find a trade-off between the stochasticity and uncontrollable velocities in the river to get as close to the waterfall as possible.

\section{Probabilistic Policy Search}
\label{sec:probabilistic_policy_search}
\begin{figure}[t]
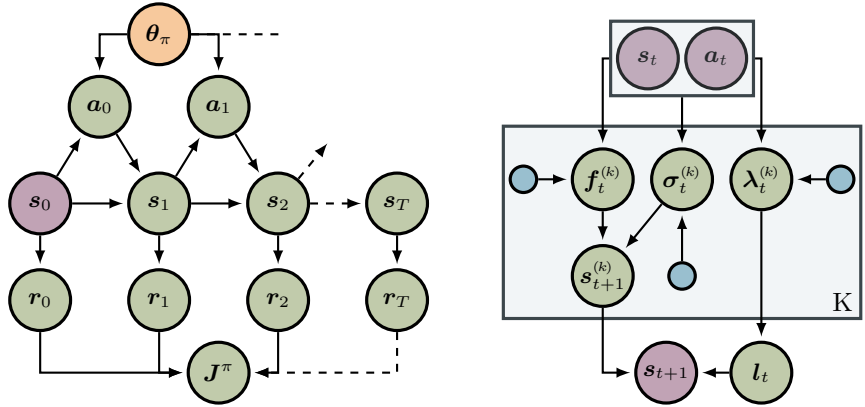

    \centering
    \begin{subfigure}[b]{.495\linewidth}
        \centering
        \includestandalone{figures/graphical_model_rl}
    \end{subfigure}
    \hfill
    \begin{subfigure}[b]{.495\linewidth}
        \centering
        \includestandalone{figures/graphical_model_mdgp}
    \end{subfigure}
    \caption{
        \label{fig:graphical_model}
        The graphical models considered in this work, where violet nodes are observed, parameters are shown in yellow and variational parameters are blue.
        The generative process for the return $J^\pi$ (left) shows how starting from $\mat{s}_0$, a trajectory of length $T$ is generated with the policy parameterized by $\mat{\theta}_\pi$.
        The return is generated by the rewards which depend on their respective states only.
        The transition model (right) separates the flow-behaviour of the river $\mat{f}_t$, the heteroscedastic noise process $\mat{\sigma}_t$ and the possibility of falling down $\mat{\lambda}_t$. 
        Latent variables $\mat{l}_t$ represent the belief that the $\nth{t}$ data point is a fall-down event.
    }
\end{figure}
We are interested in finding a policy specified by the parameters $\mat{\theta}_\pi$ which maximizes the discounted return
$J^\pi(\mat{\theta}_\pi) = \sum_{t=0}^T \gamma^t \Fun*{r}{\mat{s}_t} = \sum_{t=0}^T \gamma^t r_t$.
Starting from an initial state $\mat{s}_0$ we generate a trajectory of states $\mat{s}_0, \ldots, \mat{s}_T$ obtained by applying the action $\mat{a}_t = \Fun*{\pi}{\mat{s}_t}$ at every time step $t$.
The next state is generated using the (latent) transition function $f$, yielding $\mat{s}_{t+1} = \Fun*{f}{\mat{s}_t, \mat{a}_t}$.

Many environments have stochastic elements, such as the random drift in the Wet-Chicken benchmark from \cref{sec:wetchicken}.
We take this stochasticity into account by interpreting the problem from a Bayesian perspective where the discounted return specifies a generative model whose graphical model is shown in \cref{fig:graphical_model}.
Because of the Markov property assumed in RL, conditional independences between the states yield a recursive definition of the state probabilities given by
\begin{align}
\begin{split}
    \Prob{\mat{s}_{t+1} \given f, \mat{\theta}_\pi} &= \int \Prob{\Fun{f}{\mat{s}_t, \mat{a}_t} \given \mat{s}_t, \mat{a}_t} \Prob{\mat{a}_t \given \mat{s}_t, \mat{\theta}_\pi} \Prob{\mat{s}_t} \diff \mat{a}_t \diff \mat{s}_t, \\
    \Prob{r_t \given \mat{\theta}_\pi} &= \int \Prob{\Fun*{r}{\mat{s}_t} \given \mat{s}_t} \Prob{\mat{s}_t \given \mat{\theta}_\pi} \diff \mat{s}_t.
\end{split}
\end{align}
With stochasticity or an uncertain transition model, the discounted return becomes uncertain and the goal can be reformulated to optimizing the expected return $\Moment*{\E}{\Fun*{J^\pi}{\mat{\theta}_\pi}} = \sum_{t=0}^T \gamma^t \Moment*{\E_{\Prob{\mat{s}_t \given \mat{\theta}_\pi}}}{r_t}$.

A model-based policy search method consists of two key parts~\cite{deisenroth_pilco_2011}:
First, a dynamics model is learned from state transition data.
Second, this dynamics model is used to learn the parameters $\mat{\theta}_\pi$ of the policy $\pi$ which maximize the expected return $\Moment*{\E}{\Fun*{J^\pi}{\mat{\theta}_\pi}}$.
We discuss both steps in the following.

\subsection{An Interpretable Transition Model}
\label{sub:mdgp}
We formulate a probabilistic transition model based on high-level knowledge about the Wet-Chicken benchmark.
Importantly, we do not formulate a specific parametric dynamics model as would be required to derive a controller.
Instead, we make assumptions on a level typically available from domain experts.

We encode that given a pair of current state and action $\mat{\hat{s}}_t = \left( \mat{s}_t, \mat{a}_t \right)$, the next state $\mat{s}_{t+1}$ is generated via the combination of three things:
the deterministic flow-behaviour of the river $\mat{f}_t$, some heteroscedastic noise process $\mat{\sigma}_t$ and the possibility of falling down $\mat{\lambda}_t$.
This prior imposes structure which allows us to explicitly state what we want to learn from the data and where we do not assume prior knowledge:
How does the river flow?
What kind of turbulences exist?
When does the canoeist fall down?
How do the actions influence the system?

We formulate a graphical model in \cref{fig:graphical_model} using the data association with GPs (DAGP) model~\cite{kaiser_data_2018}, which allows us to handle the multi-modality introduced by falling down the waterfall.
We specify this separation via the marginal likelihood
\begin{align}
\begin{split}
    \label{eq:true_marginal_likelihood}
    &\Prob*{\mat{s}_{t+1} \given \mat{\hat{s}}_t} =
    \int
    \Prob*{\mat{s}_{t+1} \given \mat{\sigma}_t, \mat{f}_t, \mat{l}_t}
    \Prob*{\mat{\sigma}_t \given \mat{\hat{s}}_t}
    \Prob*{\mat{f}_t \given \mat{\hat{s}}_t}
    \Prob*{\mat{l}_t \given \mat{\hat{s}}_t}
    \diff \mat{\sigma}_t \diff \mat{l}_t \diff \mat{f}_t, \\
    &\Prob*{\mat{s}_{t+1} \given \mat{\sigma}_t, \mat{f}_t, \mat{l}_t} =
    \prod_{k=1}^K
    \Gaussian*{\mat{s}_{t+1} \given \mat{f}_t^{\pix{k}}, \left(\mat{\sigma}_t^{\pix{k}}\right)^2}_{^{\displaystyle,}}^{\Fun{\Ind}{l_t^{\pix{k}} = 1}}
\end{split}
\end{align}
where $\mat{f}_t = \left( \mat{f}_t^{\pix{1}}, \dots, \mat{f}_t^{\pix{K}} \right)$ and
$\Prob*{\mat{l}_t \given \mat{\hat{s}}_t} = \int \Multinomial*{\mat{l}_t \given \Fun{\softmax}{\mat{\lambda}_t}} \Prob*{\mat{\lambda}_t \given \mat{\hat{s}}_t} \diff \rv{\lambda}_t$
with $\Multi$ denoting a multinomial distribution.
In our case, we use $K = 2$ modes, one for staying in the river and one for falling down the waterfall.
For every data point we infer a posterior belief $\Prob{\mat{l}_t}$ about which mode the data point belongs to as we assume this separation can not be predetermined using expert knowledge.
We place independent GP priors on the $\mat{f}^{\pix{k}}$, $\mat{\sigma}^{\pix{k}}$ and $\mat{\lambda}^{\pix{k}}$.

We approximate the exact posterior via a factorized variational distribution
$\Variat*{\mat{f}, \mat{\lambda}, \mat{\sigma}, \mat{U}} = \prod_{k=1}^K\prod_{t=1}^T \Variat{\mat{f}_t^{\pix{k}}, \mat{u}^{\pix{k}}} \Variat{\mat{\lambda}_t^{\pix{k}}, \mat{u_\lambda}^{\pix{k}}} \Variat{\mat{\sigma}_t^{\pix{k}}, \mat{u_\sigma}^{\pix{k}}}$
which introduces variational inducing inputs and outputs $\mat{U}$ as described in~\cite{hensman_scalable_2015,kaiser_data_2018}.
The variational parameters are optimized by minimizing a lower bound to the marginal likelihood which can be efficiently computed via sampling and enables stochastic optimization.
For details we refer to~\cite{kaiser_data_2018}.
We obtain an explicit representation of the GP posteriors during variational inference which allows us to efficiently propagate samples through the model to simulate trajectories used for policy search.

\subsection{Policy Learning}
\label{sub:policy}
After training a transition model, we use the variational posterior $\Variat{\mat{s}_{t+1} \given \mat{\hat{s}}_t}$ to train a policy by sampling roll-outs and optimizing policy parameters via stochastic gradient descent on the expected return $\Moment*{\E}{\Fun*{J^\pi}{\mat{\theta}_\pi}}$.
The expected return is approximated using the variational posterior given by
\begin{align}
\begin{split}
    \label{eq:policy_training}
    \Moment*{\E}{\Fun*{J^\pi}{\mat{\theta}_\pi}}
    &= \sum_{t=0}^T \gamma^t \Moment*{\E_{\Prob{\mat{s}_t \given \mat{\theta}_\pi}}}{\mat{r}_t}
    \approx \sum_{t=0}^T \gamma^t \Moment*{\E_{\Variat{\mat{s}_t \given \mat{\theta}_\pi}}}{\mat{r}_t} \\
    &= \int \sum_{t=0}^T \Bigg[ \gamma^t \Moment*{\E_{\Variat{\mat{s}_t \given \mat{\theta}_\pi}}}{\mat{r}_t} \Bigg] \Prob{\mat{s_0}} \prod_{t=0}^{T-1} \Variat{\mat{s}_{t+1} \given \mat{s}_t, \mat{\theta}_\pi} \diff \mat{s}_0 \dots \diff\mat{s}_T \\
    &\approx \frac{1}{P} \sum_{p=1}^P \sum_{t=0}^T \gamma^t r_t^p.
\end{split}
\end{align}
We expand the expectation to explicitly show the marginalization of the states in the trajectory.
Due to the Markovian property of the transition dynamics, the integral factorizes along $t$.
The integral is approximated by averaging over $P$ samples propagated through the model starting from a known distribution of initial states $\Prob{\mat{s}_0}$.
State transitions can efficiently be sampled from the variational posterior of the dynamics model by repeatedly taking independent samples of the different GPs.

The expected return in \cref{eq:policy_training} can be optimized using stochastic gradient descent via the gradients $\nabla_{\theta_\pi} \Fun*{J^\pi}{\theta_\pi} \approx \frac{1}{P} \sum_{p=1}^P \sum_{t=0}^T \gamma^t \nabla_{\theta_\pi} r_t^p$ of the Monte Carlo approximation as they are an unbiased estimator of the true gradient.
The gradients of the samples can be obtained using automatic differentiation tools such as TensorFlow \cite{tensorflow2015-whitepaper}.
The $P$ roll-outs can trivially be parallelized.
Importantly, we only need a small number of Monte Carlo samples at every iteration, since we use the gradients of the samples directly.

\section{Results}
\label{sec:results}
\begin{figure}[t]
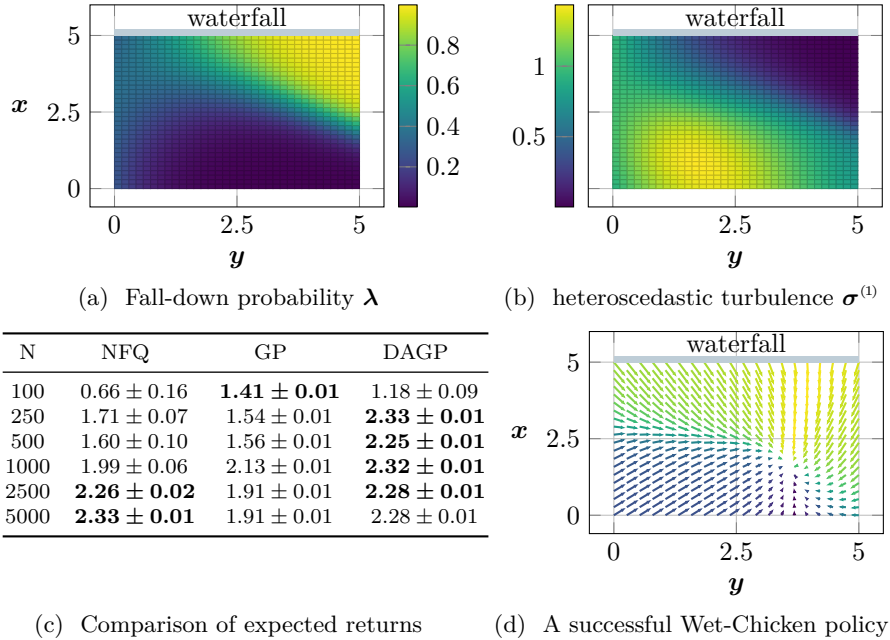

    \centering
    \begin{subfigure}{.495\linewidth}
        \centering
        \includestandalone{figures/falldown_probabilities}
        \caption{
            \label{fig:wetchicken:falldown}
            Fall-down probability $\mat{\lambda}$
        }
    \end{subfigure}
    \hfill
    \begin{subfigure}{.495\linewidth}
        \centering
        \includestandalone{figures/hetero_noise}
        \caption{
            \label{fig:wetchicken:hetero}
            heteroscedastic turbulence $\mat{\sigma}^{\pix{1}}$
        }
    \end{subfigure}
    \\[.45\baselineskip]
    \begin{subfigure}[b]{.495\linewidth}
        \centering
        \sisetup{
            table-format=-1.2(2),
            table-number-alignment=center,
            separate-uncertainty,
            table-figures-uncertainty=1,
            detect-weight,
        }
        \newcolumntype{H}{>{\setbox0=\hbox\bgroup}c<{\egroup}@{}}
        \footnotesize
        \setlength{\tabcolsep}{1pt}
        \begin{tabular}{cSSS}
            \toprule
            {N} & {NFQ} & {GP} & {DAGP} \\
            \midrule
            100 & 0.66 \pm 0.16 & \bfseries 1.41 \pm 0.01 & 1.18 \pm 0.09 \\
            250 & 1.71 \pm 0.07 & 1.54 \pm 0.01 & \bfseries 2.33 \pm 0.01 \\
            500 & 1.60 \pm 0.10 & 1.56 \pm 0.01 & \bfseries 2.25 \pm 0.01 \\
            1000 & 1.99 \pm 0.06 & 2.13 \pm 0.01 & \bfseries 2.32 \pm 0.01 \\
            2500 & \bfseries 2.26 \pm 0.02 & 1.91 \pm 0.01 & \bfseries 2.28 \pm 0.01 \\
            5000 & \bfseries 2.33 \pm 0.01 & 1.91 \pm 0.01 & 2.28 \pm 0.01 \\
            \bottomrule
        \end{tabular}
        \vspace{7ex}
        \caption{
            \label{fig:wetchicken:table}
            Comparison of expected returns
        }
    \end{subfigure}
    \begin{subfigure}[b]{.495\linewidth}
        \centering
        \includestandalone{figures/policy_quiver}
        \caption{
            \label{fig:wetchicken:policy}
            A successful Wet-Chicken policy
        }
    \end{subfigure}
    \caption{
        \label{fig:wetchicken}
        The separation of different aspects of the Wet-Chicken benchmark yields interpretable information about the probability to fall down the waterfall and the turbulence intensity.
        Successful policies can be learned based on 250 observations, while about 2500 observations are needed for NFQ.
    }
\end{figure}
To solve the Wet-Chicken problem, we first train the dynamics model on batch data sampled from the true dynamics.
The benchmark has a two-dimensional state and action spaces from which we sample uniform random transitions with varying $N$ in the range \numrange{100}{5000}.
With $N \geq 250$, our model is able to identify the underlying dynamics.
\Cref{fig:wetchicken:falldown,fig:wetchicken:hetero} show how the model has successfully identified the probabilities of falling down the waterfall and the amplitude of turbulence, both with respect to the action $(0, 0)$.
We are presented with easily separable posterior belief about different aspects of the Wet-Chicken benchmark.
This belief can be reasoned about with experts to evaluate the training result.

Next, we train a neural policy.
We sample initial states from the training data, use a horizon of $T = 5$ steps and average over $P = 20$ samples with $\gamma = 0.9$.
We use a two-layer neural network with 20 ReLU-activated units each as our policy parametrization.
\Cref{fig:wetchicken:policy} shows an example policy with a trade-off between the unpredictability on the left and the uncontrollable speed on the right.

In Table~\ref{fig:wetchicken:table}, we show expected returns averaged over 10 experiments with standard errors.
Applying random actions yields a return of about \num{1.5} and a return above \num{2.2} indicates that a proper trade-off has been found.
We compare our method to a standard GP as the dynamics model and to the model-free NFQ~\cite{riedmiller_neural_2005} trained for 20 full model learning and sampling iterations using a neural network with one 10-unit hidden layer with sigmoid activations.
The GP cannot model heteroscedastic noise or multi-modality.
It does not represent the dynamics well enough to derive a policy, illustrating our need for a more structured model.
Given enough data, NFQ is able to find successful policies.
However, our method requires about an order of magnitude less data, due to the high-level prior knowledge incorporated via the dynamics model.

\section{Conclusion}
\label{sec:conclusion}
In this paper, we demonstrated how expert knowledge can be incorporated in probabilistic policy search by imposing Bayesian structure on the learning problem.
We derived an efficient inference scheme and showed how our approach can solve the Wet-Chicken benchmark, yielding human-interpretable insights about the underlying dynamics and significantly increasing data efficiency.

\renewcommand*{\bibfont}{\footnotesize}
\printbibliography

\end{document}